\RequirePackage{fix-cm}
\documentclass[smallextended]{svjour3}       
\smartqed  
\usepackage{graphicx}

\usepackage{graphicx}
\usepackage{amsmath}
\usepackage{amssymb}

\usepackage{algorithmic}
\usepackage{algorithm}
\usepackage{subfigure}
\usepackage{array}
\usepackage{multirow,tabularx}
\usepackage{lineno}
\usepackage{pdfsync}
\usepackage{multicol}
\usepackage{color}

\newcommand{\bfs}{{\textbf{s}}}
\newcommand{\bff}{{\textbf{f}}}

\newcommand{\bfd}{{\textbf{d}}}

\newcommand{\bfx}{{\textbf{x}}}

\newcommand{\bfw}{{\textbf{w}}}
\newcommand{\bfy}{{\textbf{y}}}

\newcommand{\bflambda}{{\boldsymbol{\lambda}}}

\begin{document}

\title{When sparse coding meets ranking: A joint framework for learning sparse codes and ranking scores}

\author{Jim Jing-Yan Wang \and Xuefeng Cui  \and Ge Yu \and Lili Guo \and Xin Gao}

\institute{Jim Jing-Yan Wang, Xuefeng Cui, and Xin Gao \at
King Abdullah University of Science and Technology (KAUST),
Computational Bioscience Research Center (CBRC),
Computer, Electrical and Mathematical Sciences and Engineering (CEMSE) Division,
Thuwal, 23955, Saudi Arabia\\
\email{jimjywang@gmail.com}, \email{xuefeng.cui@kaust.edu.sa}, \email{xin.gao@kaust.edu.sa}\\
All correspondence should be addressed to Xin Gao. E-mail: xin.gao@kaust.edu.sa. Tel.: +966-12-8080323.
\and
Ge Yu and Lili Guo\at
Key Laboratory of Space Utilization, Technology and Engineering Center for Space Utilization, Chinese Academy of Sciences, Beijing, 100094, China
}

\date{Received: date / Accepted: date}

\maketitle

\begin{abstract}
Sparse coding, which represents a data point as a sparse reconstruction code with regard to a dictionary, has been a popular data representation method. Meanwhile, in database retrieval problems, learning the ranking scores from data points plays an important role. Up to now,  these two problems have always been considered separately,  assuming that data coding and ranking are two independent and irrelevant problems. However, is there any internal relationship between sparse coding and ranking score learning? If yes, how to explore and make use of  this internal relationship? In this paper, we try to answer these questions by developing the first joint sparse coding and ranking score learning algorithm. To explore the local distribution in the sparse code space, and also to bridge coding and ranking problems, we assume that in the neighborhood of each data point, the ranking scores can be approximated from the corresponding sparse codes by a local linear function. By considering the local approximation error of ranking scores, the reconstruction error and sparsity of sparse coding, and the query information provided by the user, we construct a unified objective function for learning of sparse codes, the dictionary and ranking scores. We further develop an iterative algorithm to solve this optimization problem.
\end{abstract}

\section{Introduction}

Sparse coding is a popular data representation method \cite{lee2006efficient}.  It tries to reconstruct a given data point as a linear combination of some basic elements in a dictionary, which are referred to as codewords. The linear combination coefficients are imposed to be sparse, e.g., most of the combination coefficients are zeros. The linear combination coefficient vector of a data point can be used as its new representation, and we call it the  sparse code due to its sparsity. Because of its ability to explore the latent part-based nature of the data, it has been widely used to represent data in pattern classification, image understanding, and database retrieval problems. Many sparse coding algorithms were proposed to learn the dictionary and sparse codes \cite{lee2006efficient,gao2010local,Wang20149,Wang20141630,AlShedivat20141665,Wang2013199}.

Meanwhile, in nearest neighbor-based classification and content-based database retrieval problems, the data points are usually ranked according to their similarity measures to the queries. The similarity measures are referred to as the ranking scores. Recently, methods to learn the ranking scores from the data points were proposed and showed their power in retrieval problems \cite{zhou2004ranking}.  By considering both the query information provided by the users and the distribution of the data points, efficient algorithms were developed to learn the ranking scores \cite{zhou2004ranking,yang2009ranking,Wang2014635,Wang2012ranking,Wang2012ProDis}.

It is possible to use both sparse coding and ranking score learning techniques to boost the performance of nearest neighbor searching. One may firstly map the data points to the sparse codes using a sparse coding algorithm, and then learn the ranking scores in the sparse code space. However, this strategy uses sparse coding and ranking methods independently, and assumes that they are two irrelevant problems. In this paper, we ask the following two questions about sparse coding and ranking score learning:

\begin{enumerate}
\item Is there any internal relationship between sparse coding and ranking score learning?
\item If yes, how can we explore it to boost both the data representation and ranking simultaneously?
\end{enumerate}
To answer these two questions, we propose to learn the sparse codes and ranking scores jointly to explore their internal relationship. Actually, in \cite{mairal2009supervised}, Mairal et al. proposed to learn sparse codes, a dictionary and a classifier jointly to explore the internal relationship between sparse coding and classification. However, up to now, there is no existing work considering  both sparse coding and ranking problems simultaneously.

To this end, we propose to perform sparse coding to all the data points and use the query information provided by the user to regularize the learning of the ranking scores. More importantly, to bridge the learning of sparse codes and ranking scores, and also to utilize the local distribution of the data points, we assume that in a local neighborhood of each data point, the ranking scores can be approximated from the sparse codes using a local linear function. By considering the reconstruction error and sparsity of the sparse coding problem, the local approximation error and the complexity of local ranking score approximation, and the query information regularization problems simultaneously, we construct a unified objective function for learning of the sparse codes, the dictionary and ranking scores.  By optimizing this objective function, sparse codes and ranking scores can regularize the learning of each other, and thus the internal relationship can be explored. An iterative algorithm is developed to optimize the objective function with regard to the sparse codes, the dictionary and ranking scores, using the alternate optimization strategy.

The rest parts of this paper are organized as follows: in Section \ref{sec:related}, we briefly introduce related works on sparse coding and ranking score learning. In Sections \ref{sec:method} and \ref{sec:extention}, we introduce the proposed joint sparse coding and ranking score learning method. In Section \ref{sec:exp}, we show the performance of the proposed algorithm on nearest neighbor retrieval problems using six benchmark data sets. In Section \ref{sec:conclusion}, the paper is concluded with some future work.

\section{Related Work}
\label{sec:related}
\vspace{-0.2cm}
Since our method is based on sparse coding, ranking score learning and local learning, we give some brief introduction to the relevant works. The most widely used sparse coding method was proposed by Lee et al. \cite{lee2006efficient}, which is based on iteratively  solving a $\ell_2$-constrained least square problem and a $\ell_1$-regularized least square problem.
The solutions can achieve a significant speedup for sparse coding.
This method ignores the local manifold structure of the distribution of the data points. To solve this issue, Gao et al. \cite{gao2010local} proposed Laplacian sparse coding (LapSc) to explore the local manifold structure of the data set, which is presented by a nearest neighbor graph, and used it to regularize the learning of sparse codes.
A nearest neighbor graph is constructed from the data points to present the local manifold structure, and the learned sparse codes of neighboring data points are imposed to be close.

In the learning to rank problems, Zhou et al. \cite{zhou2004ranking} also used a nearest neighbor graph to regularize the learning of ranking scores.
A disadvantage of using nearest neighbor graphs is that the ranking performance is usually sensitive to the graph parameters. To solve this problem, Yang et al. \cite{yang2009ranking} proposed a local regression and global alignment (LRGA) algorithm to use a local linear function to predict the ranking scores from the original data points in the neighborhood of each data point to explore the local manifold information.

Although local manifold information has been utilized to improve the learning of both sparse codes and ranking scores \cite{gao2010local,zhou2004ranking,yang2009ranking}, it is still not clear if there is any connection between the sparse codes and the ranking scores in a local manifold context. In this paper, we will try to predict ranking scores from the sparse codes in the local neighborhood of each data point to explore such a connection.

\section{Joint Learning of Sparse Coding and Ranking}
\label{sec:method}
\vspace{-0.2cm}
In this section, we will introduce the proposed unified sparse coding and ranking score learning method.

\subsection{Problem formulation}

Assume we have a data set of $n$ data points, denoted as $\mathcal{X}= \{\bfx_i\}_{i=1}^n$, where $\bfx_i\in \mathbb{R}^d$ is a $d$-dimensional feature vector of the $i$-th data point. In this data set, one point is provided by the user, which is named as the query, while the remaining data points are from a given database.
To indicate the query data point, we define a query indicator vector $\bflambda = [\lambda_1,\cdots,\lambda_n] \in \{1,0\}^n$, where $\lambda_i = 1$ if $\bfx_i$ is a query, and $0$ otherwise.
The problem of data retrieval is to return some data points from the database which are the most similar to the query. To this end, ranking scores are learned for the data points as similarities to the query so that the data points can be ranked according to the ranking scores, and the top ranked data points are returned as the retrieval results. The ranking scores of the data points in $\mathcal{X}$ are organized in a ranking score vector $\bff=[f_1,\cdots,f_n] \in \mathbb{R}^n$, where $f_i$ is the ranking score for the $i$-th data point. To learn the ranking score, we represent the data points as sparse codes of a dictionary first, and meanwhile learn the ranking scores from the sparse codes and query information. The following problems are considered to construct a unified objective function to learn both the sparse codes and the ranking scores.

\begin{itemize}
\item \textbf{Sparse coding} The sparse coding problem aims to learn a dictionary with $m$ codewords $\{\bfd_l\}_{l=1}^m$, and reconstruct a data point $\bfx_i$ as a sparse linear combination of the codewords,
\begin{equation}
\begin{aligned}
\bfx_i \approx \sum_{l=1}^m \bfd_l s_{il} = D \bfs_i,
\end{aligned}
\end{equation}
where $D=[\bfd_1,\cdots,\bfd_m] \in \mathbb{R}^{d\times m}$ is the dictionary matrix, $\bfd_l \in \mathbb{R}^d$ is the $l$-th codeword, and $\bfs_i = [s_{i1},\cdots,s_{im}]^\top \in \mathbb{R}^m$ is the sparse code of $\bfx_i$. To learn the dictionary and the sparse codes of the data points, the following minimization problem is considered,
\begin{equation}
\label{equ:sparsecoding}
\begin{aligned}
\underset{D, \{\bfs_i\}|_{i=1}^n}{\min}
~&
\sum_{i=1}^n \left (\left \| \bfx_i - D \bfs_i \right \|_2^2 + \alpha \|\bfs_i\|_1 \right ),\\
s.t.~& \|\bfd_l\|_2^2 \leq C, l=1,\cdots,m,
\end{aligned}
\end{equation}
where $\| \bfx_i - D \bfs_i  \|_2^2$ is the reconstruction error of the $i$-th data point measured by the squared $\ell_2$-norm distance, $\|\bfs_i\|_1$ is a $\ell_1$-norm based sparsity measure of the sparse code $\bfs_i$, and $\alpha$ is a tradeoff parameter.
By solving this problem, the data points are represented as the corresponding sparse codes. We will use the sparse codes to predict their ranking scores.

\item \textbf{Local ranking score learning} To unitize the local structure of the sparse code space, we propose to learn a local linear function for the neighborhood of each data point to approximate the ranking scores. The set of the $k$-nearest neighboring data points of $\bfx_i$ is denoted as $\mathcal{N}_i$. We propose to learn a linear function $h_i(\bfs_j)$ to approximate the ranking scores $f_j|_{j:\bfx_j \in \mathcal{N}_i}$ of data points in this neighborhood from their sparse codes $\bfs_j|_{j:\bfx_j \in \mathcal{N}_i}$,
\begin{equation}
\begin{aligned}
f_j \approx h_i(\bfs_j) =  \bfw_i^\top \bfs_j,
\end{aligned}
\end{equation}
where $\bfw_i \in \mathbb{R}^m$ is the parameter vector of the linear function of the $\mathcal{N}_i$. To learn $\bfw_i$, we propose the following minimization problem for each $\mathcal{N}_i$,
\begin{equation}
\begin{aligned}
\underset{\{\bfs_j,f_j\}_{j:\bfx_j \in \mathcal{N}_i},\bfw_i}{\min}
~& \left (
\sum_{j:\bfx_j \in \mathcal{N}_i} \left \| f_j - \bfw_i^\top \bfs_j \right \|_2^2 + \beta \|\bfw_i\|_2^2 \right ),
\end{aligned}
\end{equation}
where $\left \| f_j - \bfw_i^\top \bfs_j \right \|_2^2$ is the approximation error of ranking scores measured by squared $\ell_2$-norm, $\|\bfw_i\|_2^2$ is a square $\ell_2$-norm based regularization term used to control the complexity of the local linear function, and $\beta$ is a tradeoff parameter. An overall problem is obtained by summing up the local minimization problems over all the data points,
\begin{equation}
\label{equ:locallearning}
\begin{aligned}
\underset{\{\bfs_i,f_i,\bfw_i\}|_{i=1}^n}{\min}
~&
\sum_{i=1}^n \left ( \sum_{j:\bfx_j \in \mathcal{N}_i} \left \| f_j - \bfw_i^\top \bfs_j \right \|_2^2 + \beta \|\bfw_i\|_2^2 \right ).
\end{aligned}
\end{equation}
Note that not only the local function parameters $\bfw_i|_{i=1}^n$ are to be solved, but also the sparse codes and ranking scores.

\item \textbf{Query regularization} To unitize the query information provided by the users, we also regularize the learning of the ranking scores with the query indicator. If a data point is a query, its ranking score should be large since it is similar to itself. Thus we define a large value constant $y$ and force the ranking scores of the queries to be close to it. The following minimization problem is obtained,
\begin{equation}
\label{equ:query}
\begin{aligned}
\underset{\{f_i\}|_{i=1}^n}{\min}
~&
\sum_{i=1}^n \|f_i - y\|_2^2 \lambda_i.
\end{aligned}
\end{equation}
In this problem, when a data point $\bfx_i$ is a query ($ \lambda_i  = 1$), we minimize the squared $\ell_2$-norm distance between its ranking score $f_i$ and the large constant value $y$.
\end{itemize}

The final optimization problem is obtained by combining the problems in (\ref{equ:sparsecoding}), (\ref{equ:locallearning}), and (\ref{equ:query}),
\begin{equation}
\label{equ:objective}
\begin{aligned}
\underset{D, \{\bfs_i,f_i,\bfw_i\}|_{i=1}^n}{\min}
~&
\left \{
\sum_{i=1}^n \left (\left \| \bfx_i - D \bfs_i \right \|_2^2 +  \alpha\|\bfs_i\|_1 \right )\right.\\
&+\gamma
\sum_{i=1}^n \left ( \sum_{j:\bfx_j \in \mathcal{N}_i} \left \| f_j - \bfw_i^\top \bfs_j \right \|_2^2 + \beta \|\bfw_i\|_2^2 \right )\\
&\left . +\delta
\sum_{i=1}^n \|f_i - y\|_2^2 \lambda_i \right \},\\
s.t.~& \|\bfd_l\|_2^2 \leq C, l=1,\cdots,m,
\end{aligned}
\end{equation}
where $\gamma$ and $\delta$ are tradeoff parameters. In this problem, we need to solve a dictionary $D$, the corresponding sparse codes $\{\bfs_i\}_{i=1}^n$, the ranking scores $\{f_i\}_{i=1}^n$, and the local linear ranking score predictor parameters $\{\bfw_i\}_{i=1}^n$ of the data points. The learning of sparse codes and ranking scores are unified in a single optimization problem, and thus the learning of them are regularized by each other. This is the critical difference between the proposed method and the traditional independent sparse coding and ranking score learning algorithms which ignore the inherent connection between them.

\subsection{Optimization}

Directly solving this problem is difficult, thus we adapt the alternate optimization strategy to solve it.  The ranking scores, sparse codes and the dictionary are updated in an iterative algorithm. In each iteration, one of them is solved while the others are fixed, then their roles are switched. The iterations are repeated until a maximum iteration number is reached.

\subsubsection{Solving ranking scores}
\label{sec:rankingsovle}

When the ranking scores $\{f_i\}_{i=1}^n$ are being solved, we fix $D$  and $\{\bfs_i\}_{i=1}^n$, remove the objective terms irrelevant to ranking scores from (\ref{equ:objective}), and obtain the following problem,
\begin{equation}
\label{equ:objective_f}
\begin{aligned}
\underset{\{f_i,\bfw_i\}|_{i=1}^n}{\min}
~&
\left \{ \gamma
\sum_{i=1}^n \left ( \sum_{j:\bfx_j \in \mathcal{N}_i} \left \| f_j - \bfw_i^\top \bfs_j \right \|_2^2 + \beta \|\bfw_i\|_2^2 \right )\right .\\
&+\delta
\sum_{i=1}^n \|f_i - y_i\|_2^2 \lambda_i\\
&
\left .= \gamma \sum_{i=1}^n  g(S_i,\bff_i, \bfw_i)  +\delta
\sum_{i=1}^n \|f_i - y_i\|_2^2 \lambda_i \right \},
\end{aligned}
\end{equation}
where $g(S_i,\bff_i, \bfw_i)
=\sum_{j:\bfx_j \in \mathcal{N}_i} \left \| f_j - \bfw_i^\top \bfs_j \right \|_2^2 + \beta \|\bfw_i\|_2^2$ is defined as the local objective for each local ranking score learning problem of $\bfx_i$.
To rewrite it in the matrix form, we define a local ranking score vector for each $\mathcal{N}_i$ as
$\bff_i = [f_{i1}\cdots,f_{ik}] \in \mathbb{R}^k$, where $f_{ij}$ is the ranking score of the $j$-th nearest neighbor point of $\bfx_i$. Similarly, we define a local sparse code matrix for each $\mathcal{N}_i$ as $S_i = [\bfs_{i1},\cdots, \bfs_{ik}] \in \mathbb{R}^{m\times k}$, where $\bfs_{ij}$ is the sparse code of the $j$-th nearest neighbor point of $\bfx_i$. In this way, we rewrite $g(S_i,\bff_i, \bfw_i)$ as
\begin{equation}
\label{equ:objective_local}
\begin{aligned}
g(S_i,\bff_i, \bfw_i)
&=\sum_{j:\bfx_j \in \mathcal{N}_i} \left \| f_j - \bfw_i^\top \bfs_j \right \|_2^2 + \beta \|\bfw_i\|_2^2\\
&= \left \| \bff_i - \bfw_i^\top S_i \right \|_2^2 + \beta \|\bfw_i\|_2^2.
\end{aligned}
\end{equation}
The objective function of (\ref{equ:objective_f}) is composed of the local objective functions of all data points, thus this local objective function is to be minimized. To minimize this local objective function, we set its partial derivative with regard to $\bfw_i$ to zero,
\begin{equation}
\label{equ:w_i}
\begin{aligned}
&\frac{\partial g}{\partial \bfw_i}
=-2 S_i \bff_i^\top +2 S_i S_i^\top \bfw_i + 2\beta\bfw_i=0,\\
&\Rightarrow
\bfw_i
= \left ( S_i S_i^\top + \beta I \right )^{-1} S_i \bff_i^\top
= \Phi_i  \bff_i^\top,
\end{aligned}
\end{equation}
where
\begin{equation}
\label{equ:Phi_i}
\begin{aligned}
\Phi_i = \left ( S_i S_i^\top + \beta I \right )^{-1} S_i\in \mathbb{R}^{m\times k}.
\end{aligned}
\end{equation}
By substituting it to (\ref{equ:objective_local}), we can eliminate $\bfw_i$ from (\ref{equ:objective_local}) and rewrite it as
\begin{equation}
\label{equ:g_i}
\begin{aligned}
g(S_i,\bff_i)
=& \left \| \bff_i - \left (\Phi_i \bff_i^\top \right )^\top S_i \right \|_2^2 + \beta \left \|\Phi_i \bff_i^\top \right \|_2^2\\
=& \left \| \bff_i  \left ( I - \Phi_i^\top S_i \right )  \right \|_2^2 + \beta \left \| \bff_i \Phi_i^\top \right \|_2^2\\
=&  \bff_i \left [ \left ( I - \Phi_i^\top S_i \right )  \left ( I - \Phi_i^\top S_i \right )^\top
+ \beta \Phi_i^\top \Phi_i \right ] \bff_i^\top\\
=&  \bff_i L_i \bff_i^\top,\\
\end{aligned}
\end{equation}
where
\begin{equation}
\label{equ:L_i}
\begin{aligned}
L_i = \left [ \left ( I - \Phi_i^\top S_i \right )  \left ( I - \Phi_i^\top S_i \right )^\top
+ \beta \Phi_i^\top \Phi_i \right ] \in R^{k\times k}
\end{aligned}
\end{equation}
is a local regularization matrix for learning $\bff_i$.

Moreover, to consider the summation  of the local objective functions of all the data points in (\ref{equ:objective_f}), we can rewrite $\bff_i$ as the product of $\bff$ and a nearest neighbor indicator matrix $H_i = \{1,0\}^{n\times k}$ for each $\mathcal{N}_i$ to indicate which data points are in $\mathcal{N}_i$. The $(j,j')$-th element of $H_i$ is defined as
\begin{equation}
\begin{aligned}
{H_i}_{jj'} =
\left\{\begin{matrix}
1 , &\bfx_j~is~the~j'-th~nearest~neighor~ of \bfx_i,\\
0 ,& otherwise.
\end{matrix}\right.
\end{aligned}
\end{equation}
Then $\bff_i$ can be rewritten as,
\begin{equation}
\label{equ:f_i}
\begin{aligned}
\bff_i = \bff H_i.
\end{aligned}
\end{equation}
Substituting both (\ref{equ:g_i}) and (\ref{equ:f_i}) to (\ref{equ:objective_f}), the first term of (\ref{equ:objective_f}) can be rewritten as
\begin{equation}
\label{equ:first_term}
\begin{aligned}
&\sum_{i=1}^n g(S_i,\bff_i)= \sum_{i=1}^n \bff_i L_i \bff_i^\top\\
&= \sum_{i=1}^n \bff H_i L_i  H_i^\top \bff^\top =  \bff \left ( \sum_{i=1}^n H_i L_i  H_i^\top \right ) \bff^\top.
\end{aligned}
\end{equation}

The second term of  (\ref{equ:objective_f}) can also be rewritten in a matrix form as,
\begin{equation}
\label{equ:second_term}
\begin{aligned}
&\sum_{i=1}^n \|f_i - y\|_2^2 \lambda_i= (\bff - \bfy) diag(\bflambda) (\bff - \bfy)^\top,
\end{aligned}
\end{equation}
where $diag(\bflambda)\in \mathbb{R}^{n\times n}$ is a diagonal matrix with its diagonal vector as $\bflambda$, and $\bfy  = [y, \cdots, y] \in \mathbb{R}^n$ is an $n$-dimensional vector with all its elements as $y$.

Finally, we substitute (\ref{equ:first_term}) and (\ref{equ:second_term}) to (\ref{equ:objective_f}) and obtain the optimization problem with regard to the ranking score vector $\bff$,
\begin{equation}
\begin{aligned}
\underset{\bff}{\min}~
&
\left \{h(\bff) =
\gamma\bff \left ( \sum_{i=1}^n H_i L_i  H_i^\top \right ) \bff^\top
\right .
\\
&
\left.
+
\delta(\bff - \bfy) diag(\bflambda) (\bff - \bfy)^\top
\vphantom{\sum_{i=1}^n}
 \right \},
\end{aligned}
\end{equation}
where $h(\bff)$ is the objective function for the problem of learning $\bff$. This problem can be easily solved by setting the partial derivative of $h(\bff)$  with regard to $\bff$ to zero,
\begin{equation}
\label{equ:f_result}
\begin{aligned}
&\frac{\partial h(\bff)}{\partial \bff}=
2\gamma\bff \left ( \sum_{i=1}^n H_i L_i  H_i^\top \right )
+
2 \delta(\bff - \bfy) diag(\bflambda) = 0\\
&\Rightarrow\bff = \delta
  \bfy  diag(\bflambda) \left [\gamma \left ( \sum_{i=1}^n H_i L_i  H_i^\top \right ) + \delta diag(\bflambda)\right ]^{-1}.
\end{aligned}
\end{equation}

\subsubsection{Solving sparse codes}
\label{sec:sparsecodesolve}

When the ranking scores $\{f_i\}_{i=1}^n$ and the dictionary $D$ are fixed, and the terms irrelevant to sparse codes are removed, the optimization problem in (\ref{equ:objective}) is reduced to,
\begin{equation}
\label{equ:objective_s}
\begin{aligned}
\underset{\{\bfs_i,\bfw_i\}|_{i=1}^n}{\min}
~&
\sum_{i=1}^n \left (\left \| \bfx_i - D \bfs_i \right \|_2^2 +  \alpha\|\bfs_i\|_1 \right )\\
&+\gamma
\sum_{i=1}^n \left ( \sum_{j:\bfx_j \in \mathcal{N}_i} \left \| f_j - \bfw_i^\top \bfs_j \right \|_2^2 + \beta \|\bfw_i\|_2^2 \right ).
\end{aligned}
\end{equation}
As indicated in (\ref{equ:w_i}), the optimal solution of $\bfw_i$ is also a function of the sparse codes of data points in $\mathcal{N}_i$. Directly solving this problem is complicated, and we choose to use an EM-like algorithm to solve it. In each iteration, $\bfw_i$ is firstly estimated using the sparse codes solved in the previous iteration, and then it is fixed when the sparse codes $\{\bfs_i\}_{i=1}^n$ are updated. Moreover, we also choose to update the sparse codes one by one. When the sparse code $\bfs_i$ is considered, the others $\{\bfs_j\}_{j:j \neq i}$ are fixed. This reduces the problem in (\ref{equ:objective_s}) to
\begin{equation}
\label{equ:sc_result}
\begin{aligned}
\underset{\bfs_i}{\min}
~&
\left \| \bfx_i - D \bfs_i \right \|_2^2 +  \alpha\|\bfs_i\|_1 \\
&+\gamma
\sum_{j:\bfx_i \in \mathcal{N}_j} \left \| f_i - \bfw_j^\top \bfs_i \right \|_2^2.
\end{aligned}
\end{equation}
This problem can be easily solved by the feature-sign search algorithm \cite{lee2006efficient}.

\subsubsection{Solving the dictionary}
\label{sec:dictionarysolve}

When ranking scores and sparse codes are fixed,  only the dictionary  $D$ is considered, and the irrelevant terms are moved, the problem in (\ref{equ:objective}) is turned to
\begin{equation}
\label{equ:dictionary_result}
\begin{aligned}
\underset{D}{\min}
~&
\sum_{i=1}^n \left \| \bfx_i - D \bfs_i \right \|_2^2, \\
s.t.~& \|\bfd_k\|_2^2 \leq C, l=1,\cdots,m.
\end{aligned}
\end{equation}
This is a typical dictionary learning problem of sparse coding, and it can be solved by the Lagrange dual method  \cite{lee2006efficient}.

\subsection{Algorithm}

Based on the optimization results, we develop an iterative algorithm, which is shown in Algorithm \ref{alg}. The iterations are repeated until it meets a maximum iteration time $T$.

\begin{algorithm}
\caption{Iterative joint sparse coding and ranking score learning algorithm.}
\label{alg}
\begin{algorithmic}
\STATE \textbf{Input}: A training set of $n$ data points $\mathcal{X}= \{\bfx_i\}_{i=1}^n$, and a query indicator vector $\bflambda$;
\STATE \textbf{Input}: Tradeoff parameters $\alpha, \beta, \gamma$ and $\delta$;

\STATE Initialize $\{\bfs_i^0\}_{i=1}^n$, $D^0$, and $t=1$;

\STATE Find the $k$ nearest neighbors $\mathcal{N}_i|_{i=1}^n$ and construct the nearest neighbor indicator matrix $H_i|_{i=1}^n$ for the data points $\bfx_i|_{i=1}^n$.

\REPEAT

\STATE
\begin{enumerate}
\item Update $\Phi_i^{t}|_{i=1}^n$ and $L_i^t|_{i=1}^n$ as in (\ref{equ:Phi_i}) and (\ref{equ:L_i}) by fixing the sparse codes as $\{\bfs_i^{t-1}\}_{i=1}^n$;

\item Update the ranking score vector $\bff^t$ as in (\ref{equ:f_result});

\item Update the local ranking score predictor parameters $\bfw_i^t|_{i=1}^n$ as in (\ref{equ:w_i}) by fixing $\Phi_i^{t}|_{i=1}^n$ and $\bff_i^t|_{i=1}^n$;

\item Update the sparse codes $\bfs_i^{t}|_{i=1}^n$  one by one by solving (\ref{equ:sc_result}) by fixing $D^{t-1}$, $\bfw_i^t|_{i=1}^n$ and $f_i^t|_{i=1}^n$;

\item Update the dictionary  $D^{t}$  by solving (\ref{equ:dictionary_result}) by fixing $\bfs_i^{t}|_{i=1}^n$;

\item $t=t+1$;

\end{enumerate}

\UNTIL{$t\geq T$}

\STATE \textbf{Output}: Ranking scores  $f_i^{t-1}|_{i=1}^n$, sparse codes $\bfs_i^{t-1}|_{i=1}^n$ and a dictionary $D^{t-1}$.

\end{algorithmic}
\end{algorithm}

\section{Off-line and on-line extensions}
\label{sec:extention}

A shortage of Algorithm \ref{alg} is its high computational complexity. To calculate a ranking vector for one single query, the dictionary and the sparse codes of all the data points are updated in each iteration. This is unacceptable for an on-line retrieval system especially when the database size is large. To solve this problem, we propose a two-step strategy  including an off-line learning procedure to learn the dictionary and sparse codes of the data points of a database, and an on-line ranking procedure to learn the sparse code of a query and its ranking score vector.

\subsection{Off-line learning of dictionary and sparse codes}

In the off-line learning procedure, we only have the database of $n$ data points $\mathcal{X}= \{\bfx_i\}_{i=1}^n$, while not knowing the query. To regularize the learning of the dictionary and the sparse codes of the data points by ranking, we randomly select some presentative data points from the data set and treat them as queries. The selected query set is denoted as $\mathcal{Q}\subset \mathcal{X}$. For each query $\bfx_q \in \mathcal{Q}$, we want to learn a ranking vector $\bff^q =[f_1^q, \cdots, f_n^q] \in \mathbb{R}^n$, where $f_i^q$ is the ranking score of query $\bfx_q$ against the $i$-th data point. To learn the sparse codes, the dictionary and the ranking score vectors of the queries, we extend (\ref{equ:objective}) to (\ref{equ:objective_off}) to consider multiple queries in $\mathcal{Q}$,

\begin{equation}
\label{equ:objective_off}
\begin{aligned}
\min
~&
\left \{
\sum_{i=1}^n \left (\left \| \bfx_i - D \bfs_i \right \|_2^2 +  \alpha\|\bfs_i\|_1 \right )\right.\\
&+ \sum_{q:\bfx_q \in \mathcal{Q}}
\left [
\gamma
\sum_{i=1}^n \left ( \sum_{j:\bfx_j \in \mathcal{N}_i} \left \| f_j^q - {\bfw_i^q}^\top \bfs_j \right \|_2^2 + \beta \|\bfw_i^q\|_2^2 \right )
\right .\\
&\left .
\left . +\delta
\sum_{i=1}^n \|f_i^q - y\|_2^2 \lambda_i^q
\right ] \right \}\\
w.r.t.~
& D, \bfs_i|_{i=1}^n,
\{f_i^q,\bfw_i^q\}|_{i,q:\bfx_i \in \mathcal{X}, \bfx_q \in \mathcal{Q}},\\
s.t.~& \|\bfd_l\|_2^2 \leq C, l=1,\cdots, m,
\end{aligned}
\end{equation}
where $\bfw_i^q \in \mathbb{R}^m$ is the parameter vector of the linear function of the $\mathcal{N}_i$ to predict ranking scores of query $\bfx_q$ from the sparse codes, and $\lambda_i^q = 1$ if $\bfx_i$ is query $\bfx_q$, and 0 otherwise. To solve this problem, we adapt a similar alternate optimization strategy as the method used to solve (\ref{equ:objective}). The sparse codes and the dictionary are solved in the same way as in Section \ref{sec:sparsecodesolve} and Section \ref{sec:dictionarysolve} respectively. The ranking score vector for each query is solved independently as in Section \ref{sec:rankingsovle}.

\subsection{On-line ranking}

In the on-line ranking procedure, given the database $\mathcal{X}$ with $n$ data points and a new query $\bfx_{n+1} \in \mathbb{R}^d$, we need to calculate an $n+1$ ranking score vector $\bff = [f_1, \cdots, f_{n+1}] \in \mathbb{R}^{n+1}$ for the query. We already have the sparse codes $\bfs_1,\cdots, \bfs_n$ for data points in $\mathcal{X}$ and the dictionary $D$ learned in the off-line procedure. Thus we only need to calculate the sparse code $\bfs_{n+1}$ of the new query data point $\bfx_{n+1}$, while fixing the sparse codes of the remaining data points. We extend (\ref{equ:objective}) to (\ref{equ:objective_online}) to consider the additional query $\bfx_{n+1}$ in the on-line retrieval procedure to learn its sparse code $\bfs_{n+1}$ and its ranking scores $f_i|_{i=1}^{n+1}$,

\begin{equation}
\label{equ:objective_online}
\begin{aligned}
\underset{\bfs_{n+1}, f_i|_{i=1}^{n+1}}{\min}
~&
\left \{
\left (\left \| \bfx_{n+1} - D \bfs_{n+1} \right \|_2^2 +  \alpha\|\bfs_{n+1}\|_1 \right )
\vphantom{\sum_{i=1}^n}
\right.\\
&+\gamma
\sum_{i=1}^{n+1} \left ( \sum_{j:\bfx_j \in \mathcal{N}_i} \left \| f_j - \bfw_i^\top \bfs_j \right \|_2^2 + \beta \|\bfw_i\|_2^2 \right )\\
&\left . +\delta
\sum_{i=1}^{n+1} \|f_i - y\|_2^2 \lambda_i \right \},
\end{aligned}
\end{equation}
where $\lambda_i =  1$ if $i = n+1$, and $0$ otherwise. This problem can also be solved with an alternate optimization strategy. In an iterative algorithm, $\bfs_{n+1}$ and $f_i|_{i=1}^{n+1}$ are updated alternately. Moreover, we also assume the $k$-nearest neighbors in $\mathcal{N}_i$ of each $\bfx_i\in \mathcal{X}$ lie within $\mathcal{X}$ while not considering $\bfx_{n+1}$. In this way, in the on-line retrieval procedure, we only need to search the nearest neighbors $\mathcal{N}_{n+1}$ of $\bfx_{n+1}$, while leaving ${\mathcal{N}_i}|_{i=1}^{n}$ fixed. Actually, when we try to solve the ranking scores as in (\ref{equ:f_result}), the local regularization matrices ${L_i}|_{i=1}^n$ for the first $n$ data points are the same as the ones calculated in the off-line learning procedure and can be fixed, and we only need to update $L_{n+1}$. When $\bfs_{n+1}$ is solved, the first $n$ local learning regularization terms can be ignored because $\bfx_{n+1}$ is not in any ${\mathcal{N}_i}|_{i=1}^{n}$, and only the regularization in $\mathcal{N}_{n+1}$ needs to be considered. Thus both the computations of $\bfs_{n+1}$ and $f_i|_{i=1}^{n+1}$ are low-cost.

\section{Experiments}
\label{sec:exp}

To evaluate the proposed algorithm, we conducted experiments on six benchmark data sets and compared it to individual sparse coding and ranking score learning algorithms, as well as their simple combinations.

\subsection{Data sets and setup}

We used the Yale face database B \cite{GeBeKr01}, the USPS handwritten digit database \cite{kaynak1995methods}, the COIL100 object image database \cite{nene1996columbia}, the glass identification data set \cite{evett1987induction}, the climate model simulation crashes data set \cite{gmdd6}, and the
ionosphere data set  \cite{sigillito1989classification}.
The statistical information of these data sets are given in Table \ref{tab:data}. To conduct the retrieval experiments, we employed the 4-fold cross validation. A data set was split to four folds randomly, and each fold was used as a query set, while the remaining three folds were combined and used as the database set. We first performed the off-line learning procedure on the database set, and then performed the on-line ranking procedure to each query in the query set. The retrieval performance of the ranking is measured by the receiver operating characteristic (ROC) curve and the recall-precision curve. The area under ROC curve (AUC) was also used as a single performance measure.

\begin{table}[!th]
\centering
\caption{Statistical information of data sets.}
\label{tab:data}
\begin{tabular}{|l|r|r|r|}
\hline
Data set & \# Data points & \# Classes & \# Features \\
\hline\hline
Yale B & 2, 414 & 38 & 1, 024 \\
\hline
USPS & 9, 298 & 10 & 256 \\
\hline
COIL100 & 7, 200 & 100 & 1, 024 \\
\hline
Glass & 214 & 6 & 10 \\
\hline
Climate  & 540 & 2 & 18 \\
\hline
Ionosphere   & 351 & 2 & 34 \\
\hline
\end{tabular}
\end{table}

\subsection{Results}

\subsubsection{Comparison against independent sparse coding and ranking methods}

\begin{figure*}[!htb]
\centering
\subfigure[Yale B]{
\includegraphics[width=0.48\textwidth]{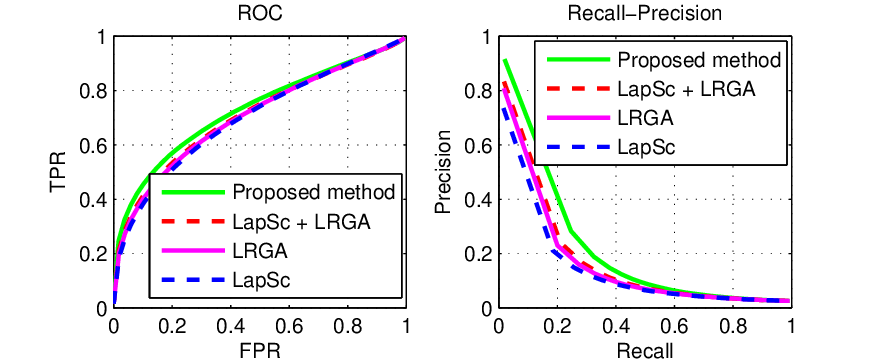}}
\subfigure[USPS]{
\label{fig:usps}
\includegraphics[width=0.48\textwidth]{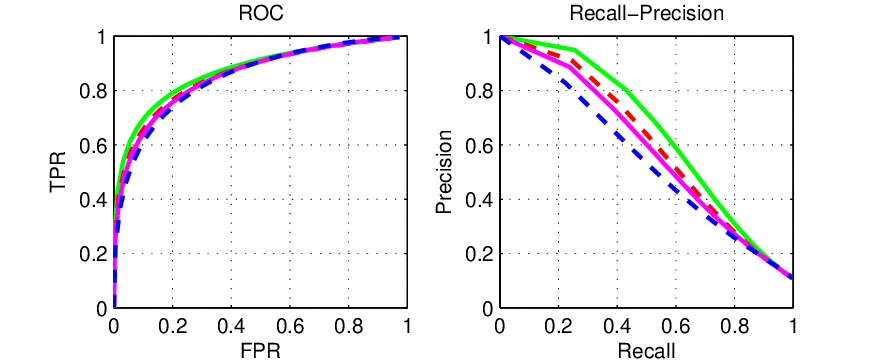}}
\subfigure[COIL100]{
\includegraphics[width=0.48\textwidth]{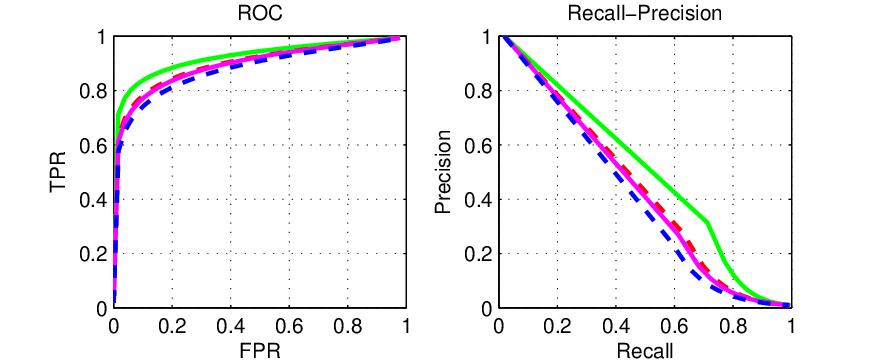}}
\subfigure[Glass]{
\includegraphics[width=0.48\textwidth]{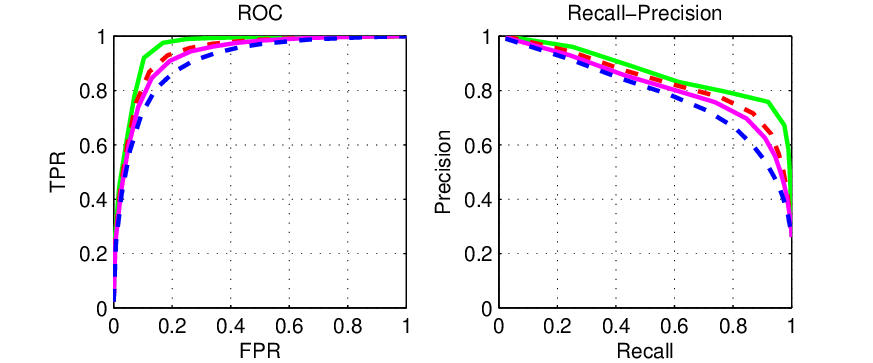}}
\subfigure[Climate]{
\includegraphics[width=0.48\textwidth]{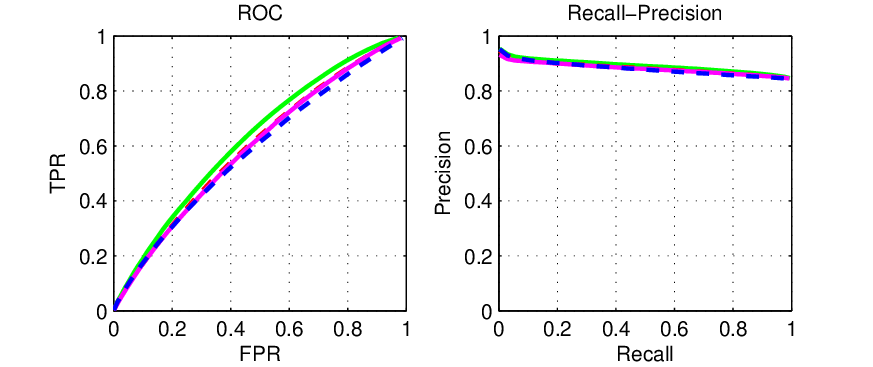}}
\subfigure[Ionosphere]{
\includegraphics[width=0.48\textwidth]{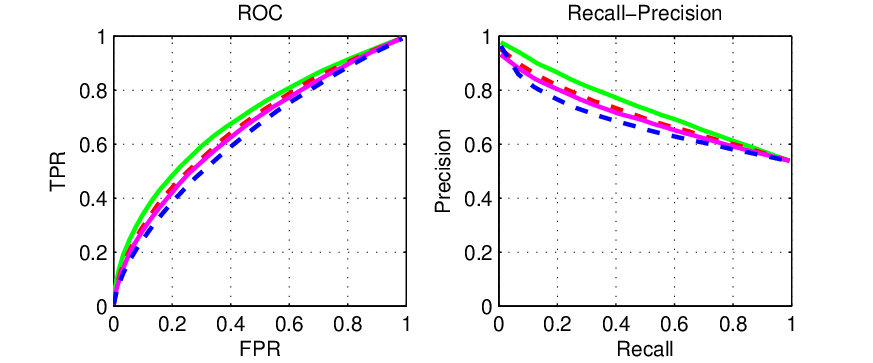}}
\\
\caption{ROC and recall-precision curves of different methods on six data sets.}
\label{fig:roc}
\end{figure*}

We compared our joint sparse coding and ranking score learning algorithm with a state-of-the-art sparse coding method, LapSc \cite{gao2010local}, and a state-of-the-art ranking score learning algorithm, LRGA \cite{yang2009ranking}, and their simple combination, i.e., using LapSc to learn sparse codes and then using the sparse codes with LRGA to learn the ranking. Both of these two individual sparse coding and ranking algorithms are based on manifold learning.
Note that we did not consider supervised sparse coding algorithms for fair comparison since the proposed algorithm is an unsupervised learning algorithm. The ROC curves of the compared methods are given in Fig. \ref{fig:roc}. From these figures, we can see that the proposed method clearly outperforms the independent sparse coding algorithm, the ranking score learning algorithm, and their simple combination on the six different data sets. In all the plots, the ROC curves of the proposed method are closer to the top-left corner of the figures than any other method, while the recall-precision curves of the proposed method are closer to the top-right corner of the figures than other methods. This indicates an overall better retrieval performance. These are strong evidences of the advantage of the joint sparse coding and ranking method over the independent sparse coding and ranking methods. This claim can be further supported by the AUC values of ROC curves in Table \ref{tab:auc}. Over the six data sets, the proposed method achieves the highest AUC values. For example, for data set COIL100, only the proposed method achieves an AUC value higher than 0.90. Moreover, it is interesting to see that LRGA outperforms LapSc in most cases, while incorporating LapSc to LRGA in a simple way does not achieve significant improvement over LRGA.
For example, in Fig. \ref{fig:usps}, the recall-precision curve of LRGA is significantly closer to the top-right corner than that of LapSc, and the recall-precision curves of LRGA and the simple combination LRGA+LapSc are close.
Although both LapSc and LRGA explore the manifold structure of the data set, LapSc applies manifold regularization in the sparse code space, while LRGA directly regularizes the ranking scores by the manifold. This means manifold learning in the representation space does not guarantee an effective ranking result from this space, and it is necessary to perform local learning to the ranking score space like LRGA. Moreover, performing LRGA in the sparse code space provided by LapSc can also improve the retrieval results, but the improvement is marginal. Only when sparse coding and ranking is performed jointly by the proposed method, significant improvements are achieved. This means sparse coding has the potential to improve the performance of ranking, but it is necessary to explore the inner relation between them.

\begin{table}[!th]
\centering
\caption{AUC values of different methods.}
\label{tab:auc}
\begin{tabular}{|l||r|r|r|r|}
\hline
Data set & Proposed method  & LapSc + LRGA & LRGA & LapSc \\
\hline\hline
Yale B & 0.7333 & 0.7130 & 0.7091 & 0.7032 \\
\hline
USPS & 0.8524 & 0.8401 & 0.8365 & 0.8293 \\
\hline
COIL100 & 0.9070 & 0.8834 & 0.8793 & 0.8637 \\
\hline
Glass & 0.9666 & 0.9492 & 0.9403 & 0.9216 \\
\hline
Climate & 0.6097 & 0.5902 & 0.5862 & 0.5821 \\
\hline
Ionosphere & 0.6946 & 0.6692 & 0.6589 & 0.6362 \\
\hline
\end{tabular}
\end{table}

\subsubsection{Sensitivity to parameters}

\begin{figure}[!htb]
\centering
\subfigure[$\alpha$]{
\label{fig:alpha}
\includegraphics[width=0.31\textwidth]{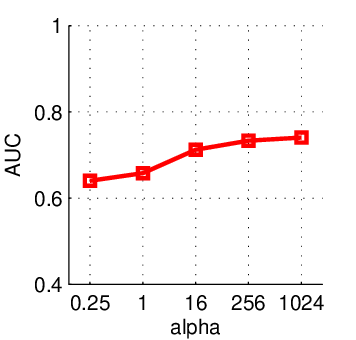}}
\subfigure[$\beta$]{
\label{fig:beta}
\includegraphics[width=0.31\textwidth]{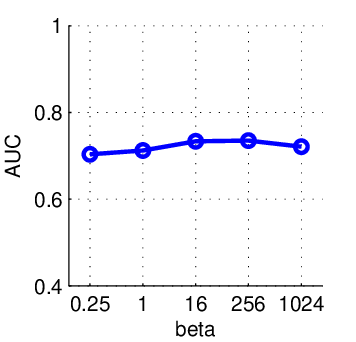}}
\subfigure[$\delta$]{
\label{fig:delta}
\includegraphics[width=0.31\textwidth]{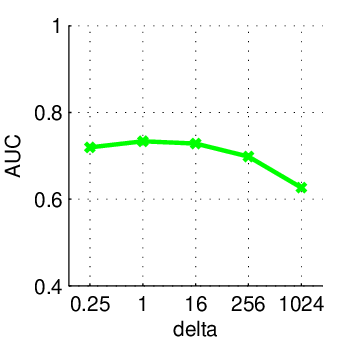}}
\caption{Parameter sensitivity curves.}
\label{fig:para}
\end{figure}

There are three tradeoff parameters $\alpha$, $\beta$ and $\gamma$ in the objective function (\ref{equ:objective}). We are also interested in the sensitivity of the proposed method to these parameters, and we plot the AUC values against different values of the parameters in Fig. \ref{fig:para}. The parameter sensitivity analysis is performed over the Yale face database B. In Fig. \ref{fig:alpha}, we can see that AUC tends to increase when $\alpha$ is increased, indicating that a sparser representation can achieve better performance. However, it seems the performance is stable when a large value of $\alpha$ is given. From Fig. \ref{fig:beta}, it can be seen that the proposed algorithm is stable to the parameter $\beta$, while from Fig. \ref{fig:delta}, it seems that a large value of $\delta$ reduces the weight of local learning and obtains a lower AUC. This indicates the importance of the local learning.

\section{Conclusion and Future Work}
\label{sec:conclusion}
\vspace{-0.2cm}
Is there any internal relationship between a popular data representation method, sparse coding, and an important procedure of the nearest neighbor search problem, ranking score learning? To answer this question, in this paper, we assume such a relationship exists, and propose to explore it by using a local linear function to approximate the ranking scores from the sparse codes in the local neighborhood of each data point. A unified objective function is constructed based on the local learning of ranking scores from sparse codes, and also based on the sparse coding and query information regularization problems. By iteratively optimizing it with regard to the sparse codes, the dictionary, and ranking scores, we develop the first joint sparse coding and ranking score learning algorithm. If the assumption holds, it is expected that the joint method which takes the advantage of this internal relationship should outperform the independent sparse coding and ranking algorithms which ignore this relationship. The proposed algorithm demonstrates superior performance over the existing sparse coding algorithm, the ranking score learning algorithm, and their simple combination. This verifies our assumption and reveals the existence of the internal relationship between the sparse coding and ranking score learning problems.

In the future, we will extend the proposed method to big data ranking, by using distributed computing models \cite{tang2016locationspark,tang2016similarity,al2016similarity}. Moreover, we will investigate more representation methods for ranking purpose besides sparse coding, such as using Bayesian networks for data representation and ranking score learning \cite{Fan20142439,Fan2014200,fan2015improved}. In the proposed model, we use a simple squared $\ell_2$-norm distance to measure the loss of ranking score learning. However, in the test process, we use the AUC as the performance measure. In the future, we will also study how to minimize a loss function that directly corresponds to AUC instead of the squared $\ell_2$-norm distance to obtain the optimal performance measure directly \cite{wang2016optimizing,Wang20151915}.

\section*{Acknowledgement}

The research reported in this publication was supported by funding from King Abdullah University of Science and Technology (KAUST) and the National Natural Science Foundation of China under the grant No. 61502463.


\end{document}